\documentclass[letterpaper]{article} 
\usepackage[preprint]{aaai2027}
\usepackage[hyphens]{url}  
\usepackage{times}  
\usepackage{helvet}  
\usepackage{courier}  
\usepackage[hyphens]{url}  
\usepackage{graphicx} 
\usepackage{natbib}  
\usepackage{caption} 
\usepackage{algorithm}
\usepackage{algorithmic}
\usepackage{amsmath}
\usepackage{multirow}
\usepackage{booktabs}
\usepackage{xcolor}
\usepackage{makecell}
\usepackage{amsfonts}
\newcommand{\pub}[1]{{\color{black}{\tiny{[{#1}]}}}}

\usepackage{newfloat}
\usepackage{listings}
\DeclareCaptionStyle{ruled}{labelfont=normalfont,labelsep=colon,strut=off} 
\floatstyle{ruled}
\newfloat{listing}{tb}{lst}{}
\floatname{listing}{Listing}
\title{Fuzzy-MoE: Interpretable Regime-Conditioned Expert Routing for Non-Stationary Multivariate Time Series Forecasting}
\author{
    Lan Guo\textsuperscript{\rm 1},
    Jie Xiao\textsuperscript{\rm 1},
    Zhao Su\textsuperscript{\rm 1},
    Jun Shen\textsuperscript{\rm 2},
    Haoran Li\textsuperscript{\rm 3},
    Weixia Ma\textsuperscript{\rm 1},
    Qingguo Zhou\textsuperscript{\rm 1},
    Binbin Yong\textsuperscript{\rm 1}\thanks{Corresponding author:
    Binbin Yong(yongbb@lzu.edu.cn)}
}

\affiliations{
    \textsuperscript{\rm 1} School of Information Science and Engineering, Lanzhou University\\
    \textsuperscript{\rm 2} School of Computing and Information Technology, University of Wollongong\\
    \textsuperscript{\rm 3} Department of Data Science and AI, Monash University\\
    
}

\usepackage{bibentry}

\begin{document}

\maketitle

\begin{abstract}
In non-stationary multivariate time series, different variables and samples often exhibit heterogeneous latent dynamic states, while existing deep forecasting models usually compress them into a unified end-to-end mapping, leading to suboptimal modeling of time-varying dynamics and limited interpretability regarding which forecasting mechanism is activated under different latent states. To overcome these limitations, we reformulate time series forecasting as a unified framework of latent temporal state identification and interpretable expert routing, and propose Fuzzy-MoE, a fuzzy logic-based dynamic Mixture-of-Experts model. Fuzzy-MoE consists of multiple parallel expert mapping networks and a dual-view fuzzy router. By jointly exploiting local convolutional dynamics and global segmented statistics, the router infers latent temporal states and computes expert activation strengths through learnable Gaussian membership functions, enabling explicit IF-THEN rule-based expert selection. This fine-grained routing strategy allows different variables within the same sequence to activate different experts, effectively capturing heterogeneous temporal dynamics while improving model interpretability. Experimental results on multiple public time series benchmark datasets show that Fuzzy-MoE significantly outperforms mainstream forecasting methods in forecasting accuracy. Moreover, fuzzy memberships and rule activations provide interpretable routing diagnostics, demonstrating the effectiveness of the proposed framework in both forecasting performance and mechanism transparency. Unlike traditional MoE models that use black-box routing, Fuzzy-MoE’s routing is based on clear, interpretable fuzzy rules. This makes the expert selection transparent and traceable.
\end{abstract}

\section{Introduction}

Non-stationary multivariate time series forecasting is fundamental to numerous real-world applications\cite{application1,application2,application3}, yet its inherent heterogeneity remains a primary obstacle. In such data, the underlying temporal dynamics often vary not only across different time segments (sample-level) but also across different variables within the same observation window (channel-level). This diversity implies that no single, unified mapping function can adequately capture all coexisting dynamic regimes; forcing all inputs into a one-size-fits-all end-to-end model inevitably leads to suboptimal forecasting performance~\cite{autoformer,dlinear,patchtst,itransformer,sde,wpmixer}, as the model fails to explicitly identify which latent state governs the current input~\cite{challenge1}.
\begin{figure}[!t]
  \centering
  \includegraphics[width=1\columnwidth]{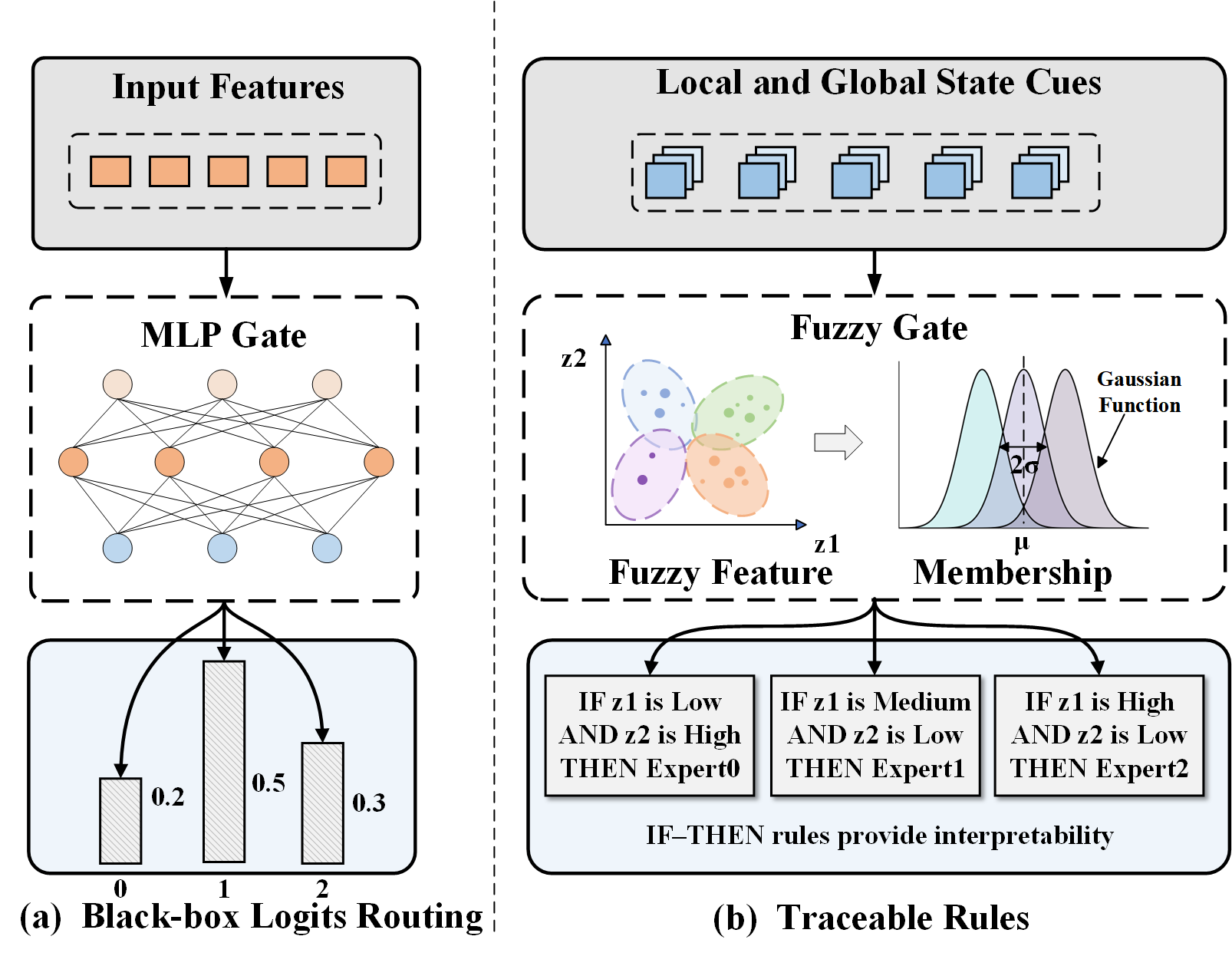}
  \caption{Comparison of classic black-box MLP gate and our interpretable fuzzy gate. (a) Traditional MLP gate generates unexplainable routing logits without traceable decision rules; (b) Our fuzzy gate fuses local and global temporal cues, computes expert weights via Gaussian fuzzy membership, and yields readable IF-THEN routing rules for fine-grained variable-wise expert assignment.}
  \label{fig:compare}
\end{figure}

To cope with this heterogeneity, recent works have turned to Mixture of Experts (MoE) architectures, which employ multiple expert networks to cover different pattern subspaces. However, the routing mechanisms in these models, which determine how inputs are assigned to experts, remaining a critical bottleneck~\cite{challenge2,challenge22}. The vast majority of existing predictors rely on black-box linear projections followed by Softmax gating, as shown in Figure~\ref{fig:compare}(a). Such designs not only lack interpretability, as the resulting weights offer no insight into which temporal features drive the selection of a particular expert, but they are also prone to gate-collapsing, where a few experts dominate all inputs while others degenerate. Consequently, despite strong fitting capabilities, these models provide no transparent rationale for their decisions, severely limiting trust and deployability in high-stakes domains. Standard MoE has a problem. It outputs routing weights but cannot explain why. It cannot answer: 'Why was this expert chosen?' This makes the model hard to trust.

We address these challenges by fundamentally reformulating the forecasting task. Instead of learning a single direct mapping, we propose a paradigm shift toward latent temporal state identification followed by interpretable expert routing. Our core philosophy is that the model should first diagnose the underlying dynamic regime of each input, and then explicitly select the expert network best suited to that regime. This decomposition transforms the opaque end-to-end forecasting into a two-stage, inspectable process: state reasoning and conditional forecasting. Our method is different from standard MoE. We do not hide routing weights inside. Instead, we create clear IF-THEN rules. You can see which expert is chosen and why.

In this work, we proposed Fuzzy-MoE, a fuzzy logic-based dynamic Mixture-of-Experts model, as shown in Figure~\ref{fig:compare}(b). It extracts complementary state cues via a dual-view router that captures local convolutional dynamics and global segmented statistics. These cues are projected into a low-dimensional fuzzy space, where routing weights are generated through learnable Gaussian membership functions and explicit IF-THEN rule firings. Crucially, the router operates at the sample-channel level, allowing different variables within the same sequence to activate different experts, a fine-grained adaptability that mirrors real-world heterogeneity. The product T-norm aggregation and temperature-scaled Softmax ensure numerical stability while preserving the physical meaning of each rule’s activation, yielding a fully traceable decision path. Each rule is simple: IF fuzzy variable is in range $[\mu - \sigma, \mu + \sigma]$, THEN activate expert$_k$. You can understand the rule without seeing the weights. In summary, our contributions are as follows:
\begin{itemize}
    \item We redefine non-stationary multivariate forecasting from the traditional sequence-to-sequence mapping to an interpretable state-conditioned expert routing problem. This formulation extends the contribution beyond a single forecasting task, pointing to a more general adaptive reasoning paradigm under heterogeneous data dynamics.
    \item We propose a transparent routing architecture that extracts local and global state cues, and generates expert weights through learnable Gaussian membership functions and explicit IF-THEN rules, making latent state identification an inspectable and interpretable intermediate process.
    \item Our model assigns experts at the sample-channel level, thereby enabling different variables within the same sequence to invoke distinct forecasting mechanisms. Furthermore, the framework produces multi-level diagnostic signals, including fuzzy memberships, expert weights, and rule activations, which collectively reveal the model's confidence, uncertainty, and the specific rationale underlying each expert selection.
\end{itemize}

\section{Related Works}
\subsection{Time Series Forecasting}
Time series forecasting is a core problem in the field of data mining and has received widespread attention for a long time~\cite{tsf1,tsf2}. Traditional methods, represented by autoregressive moving average (ARIMA) perform well when dealing with stationary linear data but lack the ability to fit nonlinear complex patterns. Machine learning methods like support vector regression (SVR) improve forecasting power through nonlinear kernels or ensemble strategies, they still rely on manually designed features and struggle to automatically uncover deep temporal representations.

In recent years, deep learning models have made significant progress in time series forecasting. Transformer-based models leverage self-attention mechanisms to capture global dependencies and have become one of the mainstream paradigms. Representative works include: PatchTST~\cite{patchtst} and iTransformer~\cite{itransformer}, which use an inverted architecture embedding independent variable sequences into the attention mechanism to better model multivariate correlations. TimesNet~\cite{timesnet} transforms 1D time series into 2D tensors via Fourier transform, capturing both intra- and inter-period variations. Autoformer~\cite{autoformer} replaces standard self-attention with an autocorrelation mechanism to explicitly model sequence periodicity. FEDformer~\cite{fedformer} introduces frequency-domain enhancement modules to improve long-term forecasting performance. However, all these models rely on a single mapping network to convert from past to future~\cite{relatedwork1,relatedwork2,relatedwork11}, making it difficult to adapt to heterogeneous dynamic states across different variables and segments in non-stationary time series.
\subsection{Explainability in Time Series Forecasting}
The black-box nature of deep models has long limited their use in high-risk areas like finance and healthcare~\cite{relatedwork2,relatedworks222,relatedworks2222}. Existing research on interpretability mainly follows two paths~\cite{relatedworks22}. The first is post-hoc explanation, such as analyzing a trained model’s behavior using methods like attention weight visualization. However, post-hoc explanations often differ from the model’s actual decisions and can even be misleading. The second path is inherently interpretable models~\cite{relatedworks22222}, which make the reasoning process transparent through self-explanatory structures.  Temporal Fusion Transformer (TFT)~\cite{tft} provides decision support via static variable encoding and interpretable multi-head attention. While these methods do offer some level of explanation, their interpretations usually stop at the attribution level of which inputs are important, without addressing the deeper question of under what conditions does the model use which forecasting mechanism.
\subsection{Mixture of Experts}
Mixture of Experts (MoE)~\cite{jiang2024mixtral} expands model capacity through the collaborative decisions of multiple expert networks, allowing different experts to handle different subsets of data. In recent years, it has shown remarkable results in large-scale pre-trained models. For example, Google’s Switch Transformer~\cite{switch_transformer} and GLaM~\cite{glam} use MoE to scale model parameters to the trillion level while keeping computational costs nearly the same.

In the field of time series forecasting, the MoE architecture has been explored preliminarily~\cite{moe}. Some works design different experts as recurrent or convolutional networks with varying receptive fields to handle multi-scale temporal patterns; others combine MoE with transformers, using sparse gating to select different attention heads or feedforward networks. However, the gating mechanism in existing MoE predictors is essentially a black-box linear projection followed by Softmax normalization~\cite{time-moe,moe2}. They fail to generate sample-channel-level fuzzy conditional rules for fine-grained variable-wise expert allocation. Pattern-specific and distribution-shift-aware expert models mitigate temporal heterogeneity by grouping samples into distinct dynamic clusters and assigning dedicated experts per cluster, but their interpretability is limited to simple cluster-pattern matching without extracting complementary local convolutional and global segmented statistical state cues or mapping multi-scale temporal signals into traceable IF-THEN routing rules, which our Fuzzy-MoE specially designs for non-stationary multivariate forecasting tasks.
\section{Proposed Method}

In this section, we present the proposed Fuzzy-MoE framework in detail. 
\begin{figure*}[hbpt]
  \centering
  \includegraphics[width=1\textwidth]{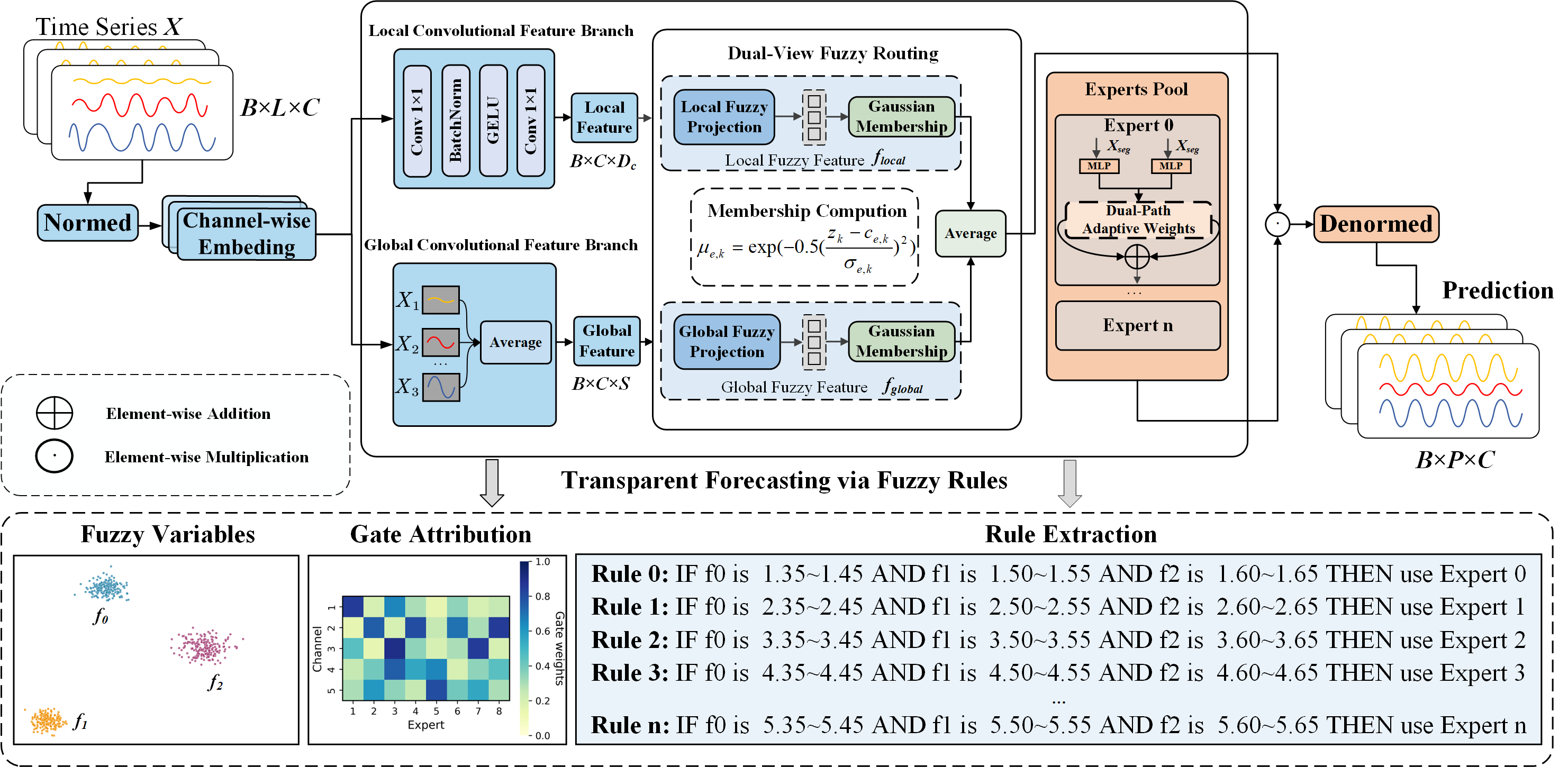}
  \caption{Overall pipeline of the proposed Fuzzy-MoE framework. It integrates input segmentation, dual-view state extraction, fuzzy routing, residual experts pool and output recovery. The dual-view fuzzy router outputs sample-channel-level expert weights, with auxiliary visualization modules to provide fuzzy embedding, gate attribution and explicit fuzzy rules for full routing interpretability.}
  \label{fig:overrall}
\end{figure*}
\subsection{Overall: From Sequence Mapping to Regime-Conditioned Routing}
\label{sec:overall}

Conventional deep forecasting models treat multivariate time series forecasting as an end-to-end sequence mapping problem: learning a unified function \(f: \mathcal{X} \mapsto \mathcal{Y}\) that directly transforms the historical observation window into the future horizon. This ``one-size-fits-all'' paradigm implicitly assumes that a single set of parameters can adequately capture all temporal dynamic patterns. However, in non-stationary time series, different variables and different segments often reside in heterogeneous dynamic regimes, rendering this assumption invalid in practice.

In this paper, we reformulate the forecasting task as a \textbf{regime-conditioned expert routing} problem. The core idea is that the model should not directly learn a single mapping; instead, it should first identify the underlying temporal regime of the current input and then select the expert network that best suits that regime to complete the forecasting. Formally, given a historical sequence \(\mathbf{X} \in \mathbb{R}^{L \times C}\) (where \(L\) is the look-back window length and \(C\) is the number of variables), the model extracts state clues from multiple views, and the fuzzy routing mechanism determines the expert selection:

\begin{equation}
\hat{\mathbf{Y}} = \sum_{k=1}^{K} \mathbf{w}_k \cdot \mathcal{E}_k(\mathbf{X}_{\text{seg}}),
\end{equation}

\noindent where \(K\) is the total number of experts, \(\mathbf{w}_k\) is the gating weight generated by the fuzzy router with explicit physical semantics, \(\mathcal{E}_k\) denotes the \(k\)-th expert network, and \(\mathbf{X}_{\text{seg}}\) is the segmented input. This formulation decomposes the forecasting process into two independently inspectable stages: \textbf{regime identification} (fuzzy routing) and \textbf{expert forecasting} (expert networks), laying the foundation for subsequent interpretability design.

\subsection{Model Architecture Overview}
\label{sec:architecture}

The overall architecture of Fuzzy-MoE consists of five core modules: input normalization and segmentation, dual-view state feature extraction, dual-view fuzzy routing, expert mapping networks, and weighted fusion with output recovery. Figure~\ref{fig:overrall} illustrates the complete model structure.

\subsubsection{Input Normalization and Segmentation}

To mitigate distribution shift, we first apply reversible instance normalization (RevIN) to the input sequence \(\mathbf{X} \in \mathbb{R}^{B \times L \times C}\) (where \(B\) is the batch size):

\begin{equation}
\tilde{\mathbf{X}} = \frac{\mathbf{X} - \mu}{\sigma + \epsilon},\quad \mu = \frac{1}{L}\sum_{t=1}^L \mathbf{X},\quad \sigma = \sqrt{\frac{1}{L}\sum_{t=1}^L \big(\mathbf{X} - \mu\big)^2}
\end{equation}

\noindent where \(\epsilon\) is a small constant for numerical stability. The normalized sequence is then divided into \(S = L / T_{\text{seg}}\) non-overlapping segments of length \(T_{\text{seg}}\), yielding \(\mathbf{X}_{\text{seg}} \in \mathbb{R}^{B \times C \times T_{\text{seg}} \times S}\). This segmentation reduces the sequence length while preserving local temporal structures, enabling efficient expert mapping at the segment level.

\subsubsection{Dual-View State Feature Extraction}

To comprehensively capture temporal state information, the model employs two complementary feature extraction pathways.

\textbf{Local convolutional dynamics pathway:} For each variable channel independently, a 1D convolutional network extracts short-term patterns and abrupt changes. This pathway consists of two convolutional layers: the first maps from 1 to 8 channels with kernel size \(k\) and stride \(s\), followed by batch normalization and GELU activation; the second is a \(1\times1\) convolution that maps back to 1 channel. The output is \(\mathbf{H}_{\text{conv}} \in \mathbb{R}^{B \times C \times D_c}\), where \(D_c\) is the convolutional output dimension.

\textbf{Global segment statistics pathway:} We compute the mean value within each segment to capture long-term trends and periodic characteristics:

\begin{equation}
\mathbf{H}_{\text{global}} = \text{mean}_{T_{\text{seg}}}(\mathbf{X}_{\text{seg}}) \in \mathbb{R}^{B \times C \times S},
\end{equation}

\noindent where \(S\) is the number of segments. These two pathways focus on different temporal scales, providing complementary state clues for subsequent fuzzy routing.

\subsection{Dual-View Fuzzy Routing Mechanism}
\label{sec:fuzzygate}

The fuzzy router is the core component that enables interpretable expert selection in Fuzzy-MoE. It consists of two independent fuzzy gate modules: the convolutional fuzzy gate \(\mathcal{G}_{\text{conv}}\) and the global fuzzy gate \(\mathcal{G}_{\text{global}}\). Each module's computation comprises three steps: fuzzy variable projection, Gaussian membership computation and rule activation, and gating weight generation. The final gating weights are obtained by fusing the outputs of the two pathways.

\subsubsection{Step 1: Fuzzy Variable Projection}

\begin{figure}[hbpt]
  \centering
  \includegraphics[width=1\columnwidth]{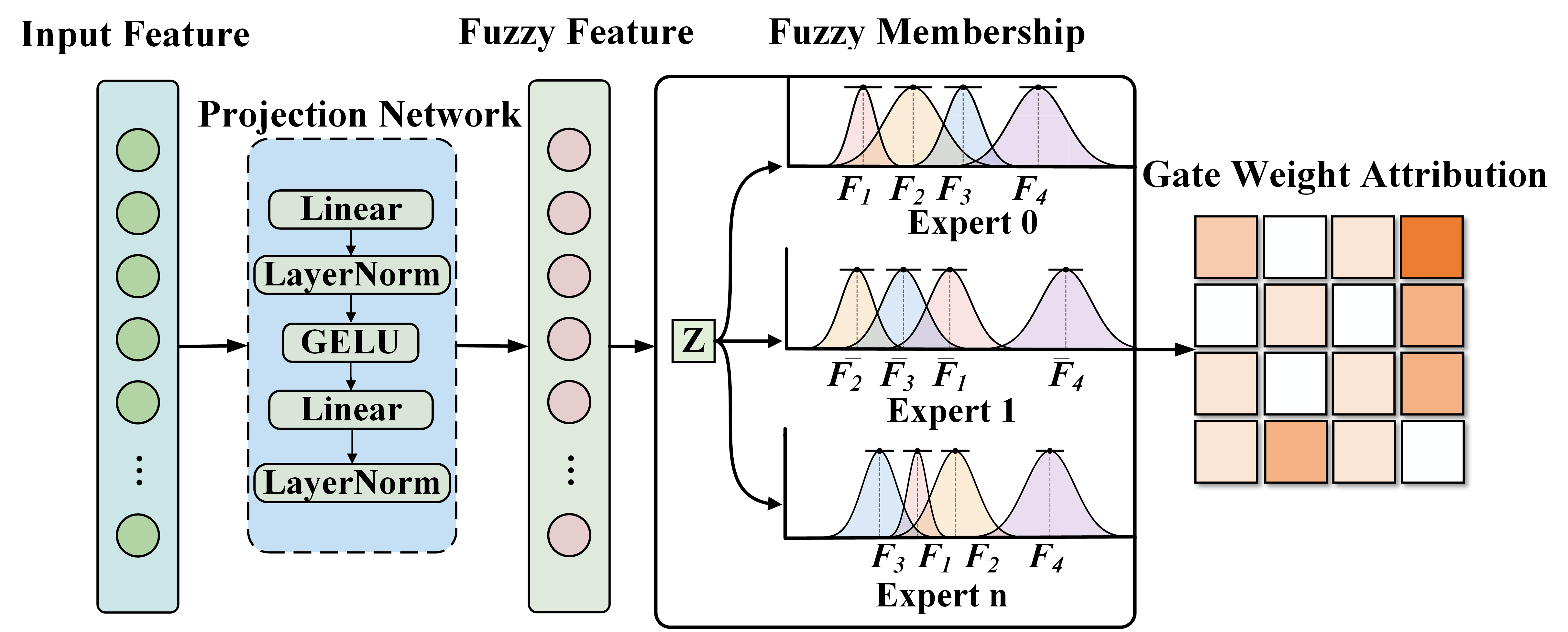}
  \caption{Detail of our Fuzzy Gate Architecture, which provides an inspectable basis for rule generation that aligns with intuitive regime-switching behaviors.}
  \label{fig:fuzzy}
\end{figure}

Given an input feature \(\mathbf{h} \in \mathbb{R}^{D}\) (where \(D\) is the feature dimension, with \(D = D_c\) for the convolutional gate and \(D = S\) for the global gate), we first project it into a low-dimensional interpretable fuzzy variable space via a multi-layer projection network:

\begin{equation}
\mathbf{z} = \text{Proj}(\mathbf{h}) \in \mathbb{R}^{M}, \quad M \ll D,
\end{equation}

\noindent where the projection network \(\text{Proj}\) consists of two linear layers with LayerNorm and GELU activation in between, which can be seen in Figure~\ref{fig:fuzzy}. \(M\) is the number of fuzzy variables. This projection maps high-dimensional abstract features into a low-dimensional latent space. By constraining the fuzzy membership functions, the model forces the network to organize these latent variables in a structured manner.

\subsubsection{Step 2: Gaussian Membership Computation and Rule Activation}

For the \(k\)-th expert, the model maintains a set of learnable Gaussian membership function parameters \(\{\boldsymbol{\mu}_k^c, \boldsymbol{\sigma}_k\}_{k=1}^K\), where \(\boldsymbol{\mu}_k^c \in \mathbb{R}^{M}\) is the center (ideal value) for each fuzzy variable, and \(\boldsymbol{\sigma}_k = \exp(\boldsymbol{\rho}_k) + \sigma_{\min} \in \mathbb{R}^{M}\) is the standard deviation (condition looseness), with \(\boldsymbol{\rho}_k\) being learnable log-standard-deviation parameters and \(\sigma_{\min}\) a minimum standard deviation constant.

The membership degree of the current sample's fuzzy variables \(\mathbf{z}\) for the \(k\)-th expert's \(m\)-th condition is computed using a Gaussian function:

\begin{equation}
u_{k,m} = \exp\left(-\frac{1}{2}\left(\frac{z_m - \mu_{k,m}^c}{\sigma_{k,m}}\right)^2\right).
\end{equation}

The value \(u_{k,m} \in (0,1]\) quantifies ``the degree to which the current sample satisfies the \(m\)-th premise condition of the \(k\)-th expert''.

Then, the firing strength of the \(k\)-th rule is obtained by aggregating the memberships of all conditions via a product T-norm:

\begin{equation}
r_k = \prod_{m=1}^{M} u_{k,m}.
\end{equation}

The physical meaning of \(r_k\) is the overall confidence that the current sample simultaneously satisfies all premise conditions of the \(k\)-th expert. The product T-norm embodies the ``AND'' operation in fuzzy logic, all conditions must be satisfied simultaneously to obtain a high firing strength.
\subsubsection{Step 3: Temperature-Scaled Softmax Gating}

Unlike existing MoE methods that apply Softmax directly to meaningless linear projections, we apply temperature-scaled Softmax normalization to the rule firing strengths, which carry explicit physical semantics:

\begin{equation}
w_k = \frac{\exp(r_k / \tau)}{\sum_{j=1}^{K} \exp(r_j / \tau)},
\end{equation}

\noindent where \(\tau\) is the temperature parameter that controls the sharpness of the gating distribution: smaller \(\tau\) yields a sharper distribution, while larger \(\tau\) leads to a smoother distribution. The key advantage of this design is that the inputs to Softmax are physically meaningful rule firing strengths rather than arbitrary linear projection values, making the generation process of gating weights traceable and interpretable.

\subsubsection{Step 4: Dual-View Fusion}

The gating computation is performed independently for each variable channel:

\begin{equation}
\mathbf{W}_{\text{conv}}^{(i)} = \mathcal{G}_{\text{conv}}(\mathbf{H}_{\text{conv}}^{(i)}) \in \mathbb{R}^{B \times K}, 
\end{equation}

\begin{equation}    
\mathbf{W}_{\text{global}}^{(i)} = \mathcal{G}_{\text{global}}(\mathbf{H}_{\text{global}}^{(i)}) \in \mathbb{R}^{B \times K}.
\end{equation}

The two gating outputs are averaged across channels and renormalized:

\begin{equation}
\tilde{\mathbf{W}} = \frac{1}{2}(\mathbf{W}_{\text{conv}} + \mathbf{W}_{\text{global}}), \quad 
\mathbf{W} = \frac{\tilde{\mathbf{W}}}{\sum_{k} \tilde{\mathbf{W}}_k + \epsilon}.
\end{equation}

The fused \(\mathbf{W} \in \mathbb{R}^{B \times C \times K}\) generates expert assignment weights independently for each variable channel at the sample-channel granularity, allowing different variables within the same sequence to select different forecasting mechanisms.

\subsection{Expert Networks and Output Fusion}
\label{sec:expert}

\subsubsection{Expert Network Design}

Each expert network \(\mathcal{E}_k\) adopts a structure with an adaptive residual connection. The trunk network is a two-layer linear mapping: \(\mathbb{R}^{S} \to \mathbb{R}^{S} \to \mathbb{R}^{T}\), where \(S\) is the number of input segments and \(T\) is the number of output segments. The shortcut branch is a single linear layer \(\mathbb{R}^{S} \to \mathbb{R}^{T}\). The final output is a weighted combination of the two branches:

\begin{equation}
\mathcal{E}_k(\mathbf{X}_{\text{seg}}) = \alpha_{\text{short}} \cdot \text{Shortcut}(\mathbf{X}_{\text{seg}}) + \alpha_{\text{trunk}} \cdot \text{Trunk}(\mathbf{X}_{\text{seg}}),
\end{equation}

\noindent where \(\boldsymbol{\alpha} = \text{Softmax}(\boldsymbol{\alpha}_{\text{raw}}) \in \mathbb{R}^2\) are learnable fusion weights. This design lets each expert balance linear and nonlinear temporal fitting.




\subsubsection{Weighted Fusion and Output Recovery}

Each expert maps the historical segments \(\mathbf{X}_{\text{seg}} \in \mathbb{R}^{B \times C \times T_{\text{seg}} \times S}\) to predicted segments \(\hat{\mathbf{Y}}_{k} \in \mathbb{R}^{B \times C \times T_{\text{seg}} \times T}\). All expert outputs are aggregated via weighted summation using the gating weights:

\begin{equation}
\hat{\mathbf{Y}} = \sum_{k=1}^{K} \mathbf{W}_k \odot \hat{\mathbf{Y}}_k \in \mathbb{R}^{B \times C \times T_{\text{seg}} \times T},
\end{equation}

\noindent where \(\odot\) denotes broadcast multiplication. The final forecasting is flattened to \(\hat{\mathbf{Y}} \in \mathbb{R}^{B \times C \times P}\) (with \(P = T_{\text{seg}} \times T\)), denormalized (multiplied by standard deviation and added back the mean), and transposed to \(\mathbb{R}^{B \times P \times C}\) as the final output.




\section{Experiments}

\subsection{Datasets and Baselines}
To validate the forecasting accuracy of our mode, we selected six commonly used datasets: 4 ETT datasetes(ETTh1, ETTh2, ETTm1, ETTm2)~\cite{informer}, Weather and Electricity for experimentation.
Based on recency, innovation and forecasting performance, we selected seven well-regarded time series models in the field as our baselines. These include both
Linear-based and Transformer-based models: WPMixer~\cite{wpmixer}, SDE~\cite{sde}, TimeMixer~\cite{timemixer}, iTransformer~\cite{sde}, Time-MoE~\cite{time-moe}, PatchTST~\cite{patchtst} and DLinear~\cite{dlinear}.
\subsection{Metrics and Implementation Details}
This experiment uses Mean Squared
Error (MSE) and Mean Absolute Error (MAE) as the evaluation metrics for the models. The dimension of fuzzy variables is set to 3. SGD optimizer with initial learning rate 2e-5 is used for updating the parameters, the training epoch is 200, the dropout rate is 0.05, and the training batch size is set to 64. All experiments were implemented with PyTorch on a single NVIDIA 3090 24GB GPU. For the details, see Appendix B \& Appendix C.
\subsection{Main Results}

Table~\ref{tab:ett_h1h2_result} compares the forecasting performance of different methods on 6 benchmark datasets.  Taking the Electricity dataset with the 720-step long forecasting horizon as an illustrative case, we compare Fuzzy-MoE against the strong MLP-based baseline iTransformer. On this challenging setting, our model achieves a MSE of 0.203 and MAE of 0.294, while iTransformer yields MSE=0.228 and MAE=0.313. Quantitatively, Fuzzy-MoE cuts MSE error by 10.96\% and MAE error by 6.07\% relative to iTransformer. 
This substantial performance gap arises because iTransformer relies on a single shared backbone to model all multivariate variables, failing to isolate distinct dynamic regimes across power load channels; by contrast, our dual-view fuzzy routing assigns specialized experts to different variables via traceable IF-THEN rules, precisely fitting heterogeneous periodic and fluctuation patterns within electricity signals under non-stationary distribution shifts.
The gains are particularly evident on challenging datasets such as Weather and Electricity, demonstrating the effectiveness of the proposed fuzzy-guided expert routing strategy in modeling complex temporal dynamics and improving forecasting accuracy.

\begin{table*}[!t]
\centering
\setlength{\tabcolsep}{3pt} 
\footnotesize
{
\begin{tabular}{l|c|cc|cc|cc|cc|cc|cc|cc|cc}
\toprule
\multicolumn{2}{c|}{Models} 
& \multicolumn{2}{c|}{Ours} 
& \multicolumn{2}{c|}{\makecell{WPMixer\\ \pub{AAAI2025}}} 
& \multicolumn{2}{c|}{\makecell{SDE\\ \pub{SIGKDD2025}}}
& \multicolumn{2}{c|}{\makecell{TimeMixer\\ \pub{ICLR2024}}}
& \multicolumn{2}{c|}{\makecell{iTransformer\\ \pub{ICLR2024}}}
& \multicolumn{2}{c|}{\makecell{Time-MoE\\ \pub{ICLR2025}}}
& \multicolumn{2}{c|}{\makecell{PatchTST\\ \pub{ICLR2023}}}
& \multicolumn{2}{c}{\makecell{Dlinear\\ \pub{AAAI2023}}} \\

\midrule
\multicolumn{2}{c|}{Metric} 
& MSE & MAE & MSE & MAE & MSE & MAE & MSE & MAE & MSE & MAE & MSE & MAE & MSE & MAE & MSE & MAE \\
\midrule
\multirow{5}{*}{ETTh1}
& 96 & \underline{0.362} & \underline{0.383} & 0.374 &0.387 & 0.387 & 0.402 & 0.381 & 0.398 & 0.394 & 0.409 & \textbf{0.352} & \textbf{0.380} & 0.376 & 0.397 & 0.396 & 0.410 \\
& 192 & \underline{0.399} & \textbf{0.406} & 0.428 &0.414 & 0.443 & 0.432 & 0.441 & 0.430 & 0.448 & 0.441 & \textbf{0.389} & \underline{0.410} & 0.426 & 0.432 & 0.445 & 0.440 \\
& 336 & \textbf{0.423} & \textbf{0.421} & 0.462 & \underline{0.437} & 0.492 & 0.457 & 0.500 & 0.459 & 0.492 & 0.465 & \underline{0.424} & 0.439 & 0.469 & 0.457 & 0.487 & 0.465 \\
& 720 & \textbf{0.433} & \textbf{0.449} & 0.482 &0.466 & 0.504 & 0.484 & 0.552 & 0.507 & 0.521 & 0.504 & \underline{0.444} & \underline{0.465} & 0.518 & 0.504 & 0.512 & 0.510 \\

\midrule
\multirow{5}{*}{ETTh2}
& 96 & \textbf{0.271} & \underline{0.334} & \underline{0.277} & \textbf{0.330} & 0.296 & 0.344 & 0.286 & 0.339 & 0.300 & 0.349 & 0.300 & 0.355 & 0.308 & 0.359 & 0.341 & 0.395 \\
& 192 & \textbf{0.332} & \textbf{0.375} & 0.351 & \underline{0.377} & 0.381 & 0.395 & 0.391 & 0.404 & 0.381 & 0.399 & \underline{0.354} & 0.383 & 0.380 & 0.406 & 0.481 & 0.479 \\
& 336 & \textbf{0.321} & \textbf{0.380} & \underline{0.363} & \underline{0.394} & 0.429 & 0.433 & 0.421 & 0.432 & 0.423 & 0.432 & 0.405 & 0.420 & 0.412 & 0.429 & 0.592 & 0.542 \\
& 720 & \textbf{0.371} & \textbf{0.420} & \underline{0.405} & \underline{0.427} & 0.435 & 0.444 & 0.468 & 0.468 & 0.426 & 0.445 & 0.465 & 0.466 & 0.435 & 0.456 & 0.840 & 0.661 \\

\midrule
\multirow{5}{*}{ETTm1}
& 96 & \textbf{0.294} & \textbf{0.343} & 0.334 & 0.368 & 0.322 & 0.363 & 0.327 & 0.364 & 0.341 & 0.376 & \underline{0.309} & \underline{0.355} & 0.323 & 0.364 & 0.345 & 0.373 \\
& 192 & \textbf{0.335} & \textbf{0.367} & 0.358 & \underline{0.375} & 0.361 & 0.385 & 0.367 & 0.386 & 0.380 & 0.394 & \underline{0.336} & 0.376 & 0.371 & 0.391 & 0.381 & 0.391 \\
& 336 & \textbf{0.362} & \textbf{0.384} & 0.384 & \underline{0.397} & 0.401 & 0.414 & 0.393 & 0.403 & 0.419 & 0.418 & \underline{0.374} & 0.405 & 0.398 & 0.408 & 0.415 & 0.415 \\
& 720 & \textbf{0.426} & \textbf{0.416} & 0.456 & 0.445 & 0.452 & 0.443 &\underline{0.451} & \underline{0.442} & 0.486 & 0.455 & 0.483 & 0.481 & 0.457 & 0.444 & 0.472 & 0.450 \\

\midrule
\multirow{5}{*}{ETTm2}
& 96 & \textbf{0.166} & \underline{0.256} & \underline{0.170} & \textbf{0.251} & 0.177 & 0.263 & 0.174 & 0.257 & 0.183 & 0.266 & 0.199 & 0.288 & 0.184 & 0.267 & 0.193 & 0.292 \\
& 192 & \textbf{0.221} & \textbf{0.292} & \underline{0.235} & \underline{0.295} & 0.248 & 0.311 & 0.236 & 0.299 & 0.252 & 0.312 & 0.248 & 0.322 & 0.246 & 0.304 & 0.284 & 0.361 \\
& 336 & \textbf{0.276} & \textbf{0.328} & \underline{0.300} & \underline{0.336} & 0.313 & 0.353 & 0.301 & 0.339 & 0.314 & 0.351 & 0.318 & 0.365 & 0.311 & 0.348 & 0.384 & 0.429 \\
& 720 & \textbf{0.368} & \textbf{0.383} & \underline{0.391} & \underline{0.392} & 0.418 & 0.415 & 0.400 & 0.400 & 0.411 & 0.406 & 0.465 & 0.451 & 0.418 & 0.414 & 0.556 & 0.523 \\

\midrule
\multirow{5}{*}{Weather}
& 96 & \textbf{0.147} & \textbf{0.198} & 0.163 & \underline{0.205} & 0.165 & 0.213 & 0.161 & 0.208 & 0.175 & 0.215 & \underline{0.158} & 0.212 & 0.175 & 0.217 & 0.196 & 0.256 \\
& 192 & \textbf{0.194} & \textbf{0.242} & \underline{0.207} & \underline{0.245} & 0.214 & 0.255 & \underline{0.207} & 0.251 & 0.225 & 0.257 & 0.211 & 0.260 & 0.220 & 0.255 & 0.238 & 0.299 \\
& 336 & \textbf{0.248} & \textbf{0.284} & 0.267 & \underline{0.291} & 0.273 & 0.297 & \underline{0.264} & 0.293 & 0.279 & 0.298 & 0.273 & 0.307 & 0.279 & 0.297 & 0.281 & 0.330 \\
& 720 & \textbf{0.315} & \textbf{0.336} & \underline{0.338} & \underline{0.337} & 0.353 & 0.352 & 0.345 & 0.345 & 0.361 & 0.350 & 0.412 & 0.400 & 0.356 & 0.348 & 0.381 & 0.381 \\

\midrule
\multirow{5}{*}{Electricity}
& 96 & \textbf{0.131} & \textbf{0.225} & 0.166 & 0.260 & \underline{0.147} & 0.245 & 0.156 & 0.247 & 0.148 & \underline{0.240} & - & - & 0.180 & 0.272 & 0.210 & 0.301 \\
& 192 & \textbf{0.145} & \textbf{0.241} & 0.175 & 0.261 & \underline{0.161} & 0.257 & 0.170 & 0.260 & 0.164 & \underline{0.256} & - & - & 0.187 & 0.279 & 0.210 & 0.304 \\
& 336 & \textbf{0.164} & \textbf{0.260} & 0.193 & 0.282 & \underline{0.176} & 0.274 & 0.187 & 0.278 & 0.177 & \underline{0.270} & - & - & 0.204 & 0.295 & 0.223 & 0.319 \\
& 720 & \textbf{0.203} & \textbf{0.294} & 0.233 & 0.314 & \underline{0.207} & \underline{0.304} & 0.227 & 0.312 & 0.228 & 0.313 & - & - & 0.245 & 0.328 & 0.257 & 0.349 \\

\bottomrule
\end{tabular}
}
\caption{Comparison of forecasting performance. The best results are highlighted in \textbf{bold} and the second-best results are
\underline{underlined}. While "-" denotes missing experimental results of the compared method on the corresponding dataset. Overall, our method consistently achieves the best performance across forecasting horizons = \{96,192,336,720\}.}
\label{tab:ett_h1h2_result}
\end{table*}

\begin{figure}[!t]
  \centering
  \includegraphics[width=1\columnwidth]{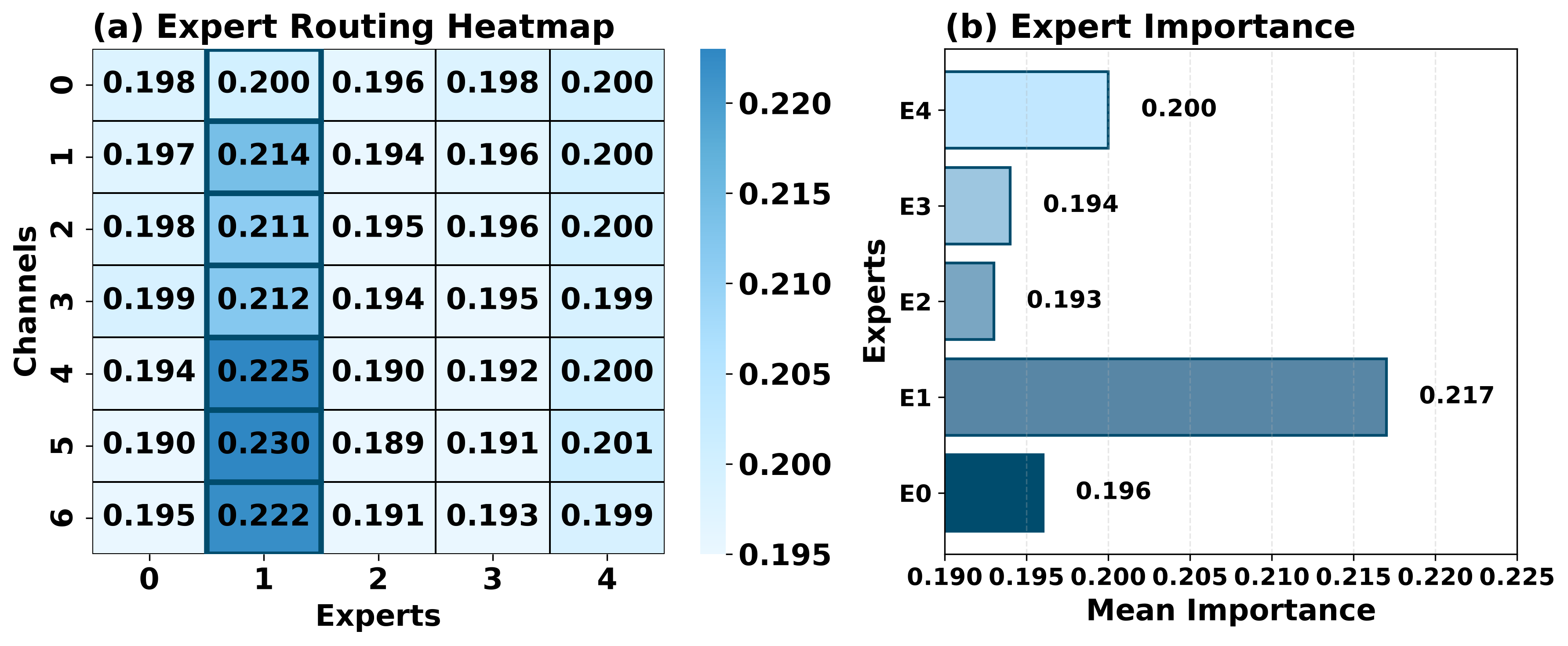}
  \caption{Visualization of expert weight attribution on the ETTh2 dataset. The routing weights are balanced yet discriminative, indicating that the fuzzy gate mechanism effectively mitigates expert collapse while encouraging expert specialization.}
  \label{fig:heatmap}
\end{figure}
Figure~\ref{fig:heatmap} visualizes the average expert routing weights learned by Fuzzy-MoE on the ETTh2 dataset.
Specifically, Expert 1 consistently receives higher routing weights across multiple channels, suggesting that it captures shared temporal dynamics, whereas the remaining experts focus on complementary latent patterns. These results demonstrate that the proposed Gaussian membership-based fuzzy routing enables adaptive and interpretable expert allocation, leading to more effective expert collaboration and improved robustness for multivariate time series forecasting.

\subsection{Ablation Study}
Table~\ref{tab:rule_abla} presents the $\Delta\text{MAE}$ values resulting from removing rules with different activation strengths on the ETTh1, ETTh2 and Weather datasets, which intuitively reflects the contribution of fuzzy rules to forecasting performance. 

\begin{table}[!t]
\centering
\small
\renewcommand{\arraystretch}{1.15}
\setlength{\tabcolsep}{8pt}
\begin{tabular}{lccc}
\toprule
Deletion Type & ETTh1 & ETTh2& Weather \\
\midrule
Top-1 activated rule & \textbf{0.012} & \textbf{0.003} 
& \textbf{0.044} 
\\
Top-2 activated rule & 0.008 & \textbf{0.003} & 0.012  \\
Lowest rule          & 0.001 & 0.000 & 0.002 \\
\bottomrule
\end{tabular}
\caption{$\Delta\text{MAE}$ results on ETTh1, ETTh2 and Weather. The best results are highlighted in \textbf{bold}.}
\label{tab:rule_abla}
\end{table}

Removing the top-1 and top-2 highly activated rules leads to substantially larger MAE increases, on ETTh1, deleting the top-1 rule yields a $\Delta\text{MAE}$ of 0.012 and removing the top-2 rule results in a $\Delta\text{MAE}$ of 0.008, while both top-1 and top-2 rule removals produce a $\Delta\text{MAE}$ of 0.003 on ETTh2. In contrast, eliminating the least activated rule only triggers negligible performance degradation, with $\Delta\text{MAE}$ values of merely 0.001 and 0.000 on the two datasets respectively. These observations demonstrate that high-confidence core IF-THEN fuzzy rules are critical for Fuzzy-MoE to achieve precise expert routing and guarantee forecasting accuracy, whereas low-activation rules exert minimal impact on the overall forecasting performance. The same trend exists on the Weather dataset. The ablation study proves the rules are real. When we remove important rules, prediction gets worse ($\Delta$MAE=0.012). When we remove unimportant rules, prediction barely changes ($\Delta$MAE=0.001). This shows the model really uses these rules. This also verifies that the Gaussian membership-based fuzzy routing rules proposed in this paper can effectively distinguish primary and secondary temporal patterns and adapt to latent temporal dynamic regimes.

We report the effect of the fuzzy temperature parameter \(\tau\) on the proposed model in Table~\ref{tab:tau}. Overall, the best performance is achieved at \(\tau\)=5, yielding the lowest MAE and MSE across all forecasting horizons.

\begin{table}[!t]
\centering
\small
\renewcommand{\arraystretch}{1.15}
\setlength{\tabcolsep}{4pt}

\resizebox{\columnwidth}{!}{
\begin{tabular}{c|c|cc|cc|cc|cc}
\toprule

\multicolumn{2}{c|}{\(\tau\)}
& \multicolumn{2}{c|}{0.3}
& \multicolumn{2}{c|}{0.8}
& \multicolumn{2}{c|}{2}
& \multicolumn{2}{c}{5}
 \\

\midrule

\multicolumn{2}{c|}{Metric}
& MSE & MAE
& MSE & MAE
& MSE & MAE
& MSE & MAE
 \\

\midrule

\multirow{4}{*}{ETTm2}
& 96  &0.167  &0.257  &\textbf{0.166}  &0.256  &0.167  &0.256  &\textbf{0.166}  &\textbf{0.255}    \\

& 192 &0.223  &0.294  &\textbf{0.221}  &0.293 
&\textbf{0.221}  &0.293  &\textbf{0.221}  &\textbf{0.292}  \\

& 336 &0.278  &0.330  &0.276  &0.328 
&0.276  &0.328  &\textbf{0.275}  &\textbf{0.327}  
\\
& 720 &0.371  &0.386  &0.369  &0.384 
&0.368  &0.383  &\textbf{0.367}  &\textbf{0.382}  
\\

\bottomrule

\end{tabular}
}
\caption{Performance comparison of different temperature \(\tau\) on ETTm2. The best results are highlighted in \textbf{bold}.}
\label{tab:tau}
\end{table}
On ETTm2, a larger temperature produces smoother fuzzy memberships and more balanced expert routing, promoting effective expert collaboration for modeling heterogeneous temporal patterns. In contrast, smaller temperature values lead to overly concentrated memberships and rigid expert assignment. 
\begin{figure}[!t]
  \centering
  \includegraphics[width=1\columnwidth]{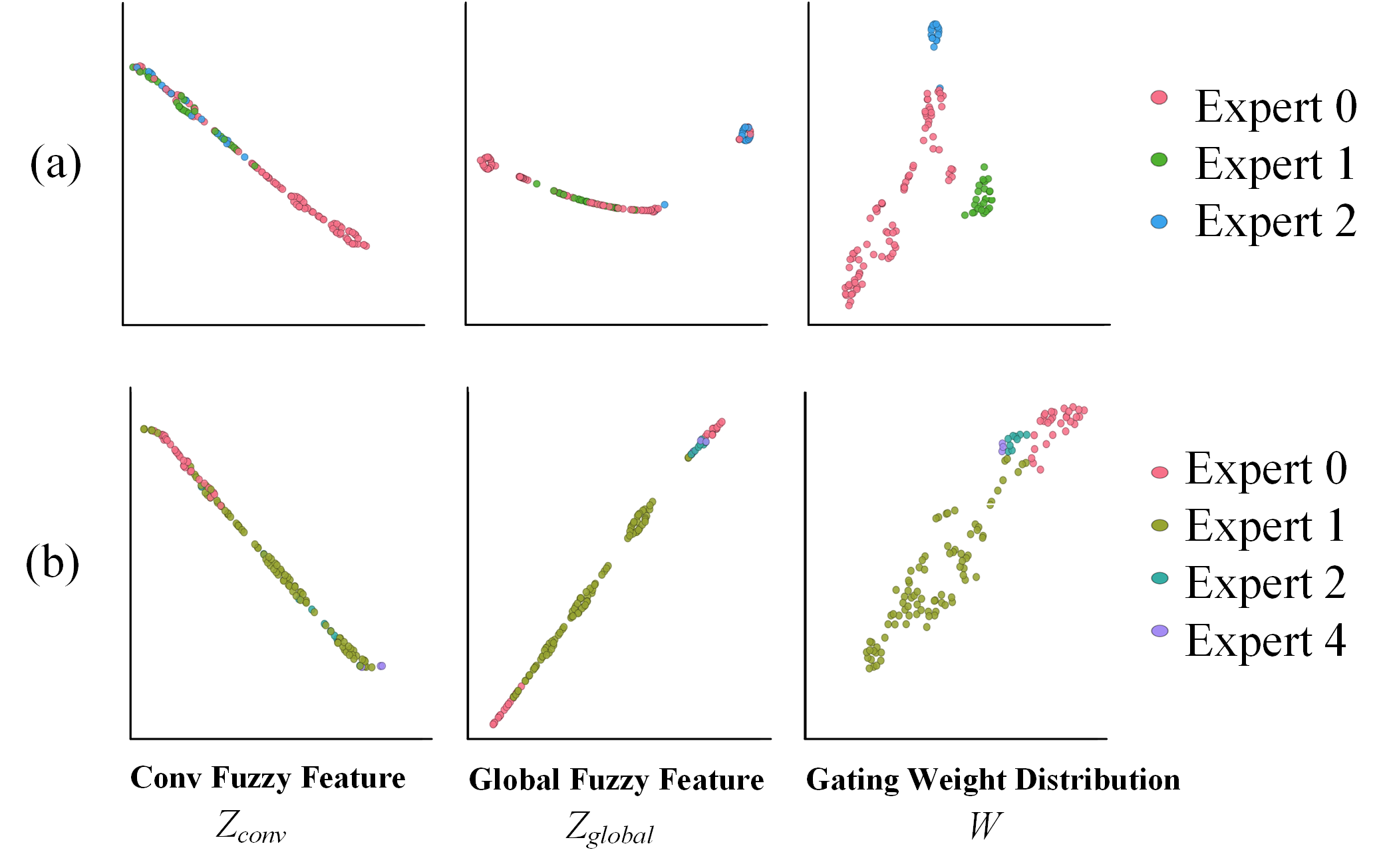}
  \newcommand{\xhdr}[1]{\vspace{1.7mm}\noindent{{\bf #1.}}}
  \caption{\xhdr{t-SNE} Visualization of Local, Global, and Fused State Representations. (a) ETTh1; (b) ETTh2.}
    
  \label{fig:tsne}
\end{figure}

Figure~\ref{fig:tsne} presents the t-SNE visualization of the feature representations learned by the local branch, global branch, and their fusion.  Compared with the individual branches, the fused representations exhibit a more structured and discriminative distribution, indicating that local and global state information provides complementary temporal cues. 
This enhanced feature separability enables the Fuzzy Gate to perform more accurate expert routing based on latent temporal states, thereby promoting expert specialization and collaboration, which ultimately contributes to the superior forecasting performance of Fuzzy-MoE.

    

\begin{table}[!t]
\centering
\small
\renewcommand{\arraystretch}{1.15}
\setlength{\tabcolsep}{4pt}

\begin{tabular}{c|c|cc|cc}
\toprule

\multicolumn{2}{c|}{Models}
& \multicolumn{2}{c|}{Fuzzy Gate}
& \multicolumn{2}{c}{MLP Gate}
 \\

\midrule

\multicolumn{2}{c|}{Metric}
& MSE & MAE
& MSE & MAE
 \\

\midrule

\multirow{4}{*}{ETTh2}
& 96  &\textbf{0.271}  &\textbf{0.334}  &0.303  &0.362  \\
& 192 &\textbf{0.332}  &\textbf{0.375}  &0.350  &0.392  \\
& 336 &\textbf{0.321}  &\textbf{0.380}  &0.331  &0.389   \\
& 720 &\textbf{0.371}  &\textbf{0.420}  &0.378  &0.426   \\

\midrule

\multirow{4}{*}{ETTm2}
& 96  &\textbf{0.166}  &\textbf{0.256}  & 0.181 & 0.263   \\
& 192 &\textbf{0.221}  &\textbf{0.292}  &0.232  &0.299    \\
& 336 &\textbf{0.276}  &\textbf{0.328}  &0.287  &0.333   \\
& 720 &\textbf{0.368}  &\textbf{0.383}  &0.376  &0.389   \\

\midrule

\multirow{4}{*}{Weather}
& 96  &\textbf{0.147}  &\textbf{0.198}  &0.149  &0.204    \\
& 192 &\textbf{0.194}  &\textbf{0.242}  &0.203  &0.250    \\
& 336 &\textbf{0.248}  &\textbf{0.284}  &0.252  &0.290   \\
& 720 &\textbf{0.315}  &\textbf{0.336}  &0.330  &0.344   \\

\bottomrule

\end{tabular}
\caption{Performance comparison of different gating mechanisms. The best results are highlighted in \textbf{bold}.}
\label{tab:gate}
\end{table}

Table~\ref{tab:gate} demonstrates that replacing the conventional MLP Gate with the proposed Fuzzy Gate consistently improves forecasting performance across different datasets and forecasting horizons. By introducing Gaussian fuzzy membership-based routing, the proposed gating mechanism provides a more discriminative and stable expert assignment than implicit MLP-based routing, leading to lower MAE and MSE in most settings. For example, when the horizon is set to 96 on ETTh2, compared with the standard MLP gate, our fuzzy gate achieves MSE of 0.271 and MAE of 0.334, corresponding to relative reductions of 10.56\% and 7.73\%, respectively. Consistent performance gains are further observed on ETTm2 and Weather.

\section{Conclusion}
This work proposes Fuzzy-MoE, an interpretable fuzzy Mixture-of-Experts framework for non-stationary multivariate forecasting. By dual-view Gaussian fuzzy routing, our model generates traceable IF-THEN rules and assigns experts per variable to resolve cross-channel heterogeneous dynamics. Sufficient benchmarks confirm our method achieves superior forecasting accuracy and intrinsic routing interpretability. For future work, we will extend the fuzzy rule extraction to online adaptive forecasting and integrate sparse expert activation to reduce computation overhead. In addition, the proposed framework offers a general, state-aware routing template that can be readily adapted to other regime-switching time series problems.

\bibliography{aaai2027}

\end{document}